# Hardware Accelerators in Autonomous Driving

Ken Power[1,2] , Shailendra Deva[1], Ting Wang[1], Julius Li[1], Ciarán Eising[2]

[1]*Motional AD Inc, USA.* [2]*University of Limerick, Ireland.*

**Abstract**

Computing platforms in autonomous vehicles record large amounts of data from many sensors, process the data through machine learning models, and make decisions to ensure the vehicle's safe operation. Fast, accurate, and reliable decision-making is critical. Traditional computer processors lack the power and flexibility needed for the perception and machine vision demands of advanced autonomous driving tasks. Hardware accelerators are special-purpose coprocessors that help autonomous vehicles meet performance requirements for higher levels of autonomy. This paper provides an overview of ML accelerators with examples of their use for machine vision in autonomous vehicles. We offer recommendations for researchers and practitioners and highlight a trajectory for ongoing and future research in this emerging field.

## 1   Introduction

Autonomous vehicles are part of the field of mobile autonomous robots as much as they are part of the automotive domain [Correll *et al.* 2022]. Autonomous taxis, also called autonomous ride-hailing, or robotaxis, are one of the significant emerging markets and opportunities for autonomous vehicles [Li *et al.* 2022]. Advances in machine learning and deep learning directly contribute to advances in vehicle autonomy. ML accelerators for autonomous driving are enabling new levels of autonomy and performance. Section 2 of this paper provides an overview of machine learning accelerators in the context of autonomous driving. This overview includes architecture styles for ML accelerators. Section 3 provides examples of how ML accelerators improve the performance of ML workloads in AVs, with a focus on machine vision use cases. Finally, this paper summarises the implications for machine vision and AV development, offers some recommendations for researchers and practitioners, and outlines a trajectory for future research in ML accelerators.

## 2   Background

Robots are autonomous when they make decisions in response to their environment rather than following pre-programmed instructions [Correll *et al.* 2022]. They achieve autonomy using multiple techniques, including signal processing, control theory, and artificial intelligence [Correll *et al.* 2022]. An autonomous vehicle is a safety-critical system with constraints on timing, scheduling, performance, and safety that are not present in other forms of robots [Jo *et al.* 2014]. The Society of Automotive Engineers (SAE) defines six levels of driving autonomation (Level 0 through Level 5) for vehicles that perform part or all of the *dynamic driving task* (DDT) on a sustained basis [SAE International 2021]. A human driver must be present in Levels 0 through 3. Level 4 (L4) and Level 5 (L5) autonomous driving systems, where there is no human driver, must operate autonomously and safely in complex and diverse conditions [SAE International 2021]. Autonomous robots need to sense and perceive their environment, which they achieve using various sensors that measures some aspects of the environment [Ben-Ari and Mondada 2018]. The most common sensors for sensing and perception are cameras, radar, and lidar.

Machine learning is increasingly adopted to enable autonomous vehicle functionality [Gharib *et al.* 2018]. In the context of autonomous driving, machine learning is a crucial tool for enabling vehicles to recognize and respond to the complex and dynamic environments in which they operate [Bachute and Subhedar 2021]. By leveraging large datasets and powerful algorithms, machine learning can help autonomous vehicles identify objects, navigate roads, and make decisions in real-time [Correll *et al.* 2022]. New machine learning models are continuously emerging and finding application in the domain of mobile robots and autonomous driving [Grigorescu *et al.* 2020]. For example, researchers have used deep learning to train autonomous vehicles to recognize and avoid obstacles, pedestrians, and other vehicles on the road [Bojarski *et al.* 2016]. The machine learning models for autonomous driving are typically trained in a public cloud or data centre. The trained models for autonomous driving deploy to and execute on computing platforms onboard the autonomous vehicles.

The capabilities of the onboard computing platform in the autonomous vehicle are a crucial factor in determining the successful performance of the machine learning models and, in turn, the vehicle's successful autonomous operation. Traditional CPUs are not sufficiently powerful to process machine learning workloads to meet the necessary performance criteria for safe and effective autonomous driving. While useful for many AI-based workloads, GPUs are just one tool in an increasingly rich ecosystem of coprocessors known as machine learning accelerators. A challenge is designing compute systems that are high performance enough to meet the demands of running increasingly sophisticated machine learning models. The challenges for L4+ systems are orders of magnitude more complex and more demanding than for L2 and L3 systems. The safety demands are more complex, with higher stakes. There are redundancy scenarios that must be factored in. Speed, performance, and latency demands are all much higher. The machine learning models are more sophisticated in L4+. There are more sensors, resulting in higher data volumes and data processing demands. 3D object detection, object tracking, and trajectory planning are autonomous driving functions that require increased compute power. ML accelerators need to deal with these challenges and offer design options that traditional compute architectures do not.

Hardware accelerators are not a new concept. Floating-point coprocessors, graphics processing units (GPUs), and video codecs are familiar and widely used examples [Patel and Hwu 2008]. What is new is the development and growth of special purpose hardware accelerators, known collectively as ML accelerators, dedicated to processing machine learning workloads [Park and Kim 2021]. [Reuther *et al.* 2022] summarize recent trends in commercial ML accelerators. Figure 1 highlights that ML accelerators exist on a continuum and can be compared along several dimensions in terms of flexibility, application, throughput, and cost. For example, at one end CPUs have general flexibility to a wide number of applications. While they can execute ML workloads, they have no specific support for them. At the other end of the continuum, neural processing units (NPU) are designed to provide increased performance for specialized ML workloads, but at the cost of being less flexible for general-purpose applications. ASICs have high parallel processing power and relatively lower cost but lack flexibility, only suitable for specific use cases. System designers therefore need to balance consider a blend of processing resources tailored to their application needs. Most of the innovations are focusing on optimising the compositions from the compute architecture spectrum for their specific applications. These accelerators are usually using silicon CMOS logic technology where the hardware manufacturing and software infrastructures are most mature and advanced to obtain the highest performance with the least costs.

There is no definitive set of standards that defines whether a processor or other hardware element is automotive grade. However, there are certain standards and operating conditions that must be met by processors, including ML accelerators, used in autonomous driving (as distinct from, say, chips used in a vehicle's entertainment system). They typically require an automotive safety integrity level (ASIL) rating in compliance with the ISO 26262 functional safety standard [ISO 2018]. Other common characteristics of automotive-qualified processors include operating temperature range from −40°C to 125°C or 150°C [Automotive Electronics Council 2019], and supporting automotive-specific communications protocols and signaling interfaces [Staron 2021]. ML silicon should also be qualified according to AEC-Q100 standard for ICs and AEC-Q104 for multi-package IC reliability standards [Automotive Electronics Council 1994]. Other system considerations include protection against electro-static discharge and support for failure modes effects and diagnostic analysis (FMEDA).

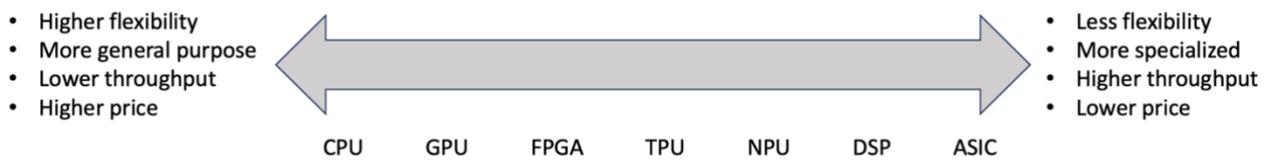

**Figure 1: Continuum of processor architectures from general purpose CPUs to special-purpose ML accelerators**

## 3   Matching Machine Vision Workloads to Processor Architectures

These engines are used mostly for core ML processing in autonomous driving machine vision tasks. Figure 2 shows the common stages of processing for the three primary sensor modalities used in AVs. Figure 2 shows some common examples of tasks processed at each stage. While state-of-the-art machine vision algorithms make use of ML, when it comes to the full machine vision pipeline a lot of work is also performed in CPUs or DSP. Hardware-based ML accelerators are most useful in performing core ML processing on camera, radar, and lidar data.

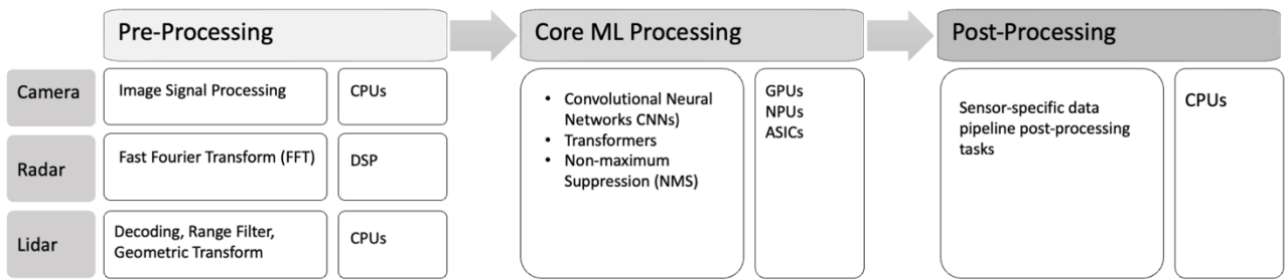

**Figure 2: Examples of matching machine vision tasks to processor types for different sensor modalities**

Support for functional safety and cybersecurity are table stakes. Design trade-offs to consider include performance, power, and cost. Factors influencing system performance include on-chip memory (e.g., HBM, SRAM), off-chip memory (e.g., DDR), and core clock rate. Moving data around is expensive and one of the main contributors to system latency. We need to consider relative strengths of interconnect options for connecting multiple chips in a system. Options include PCIe, MIPI, Ethernet, and GMSL. Special bus structures, non-uniform memory access, and custom cache coherence mechanisms can improve performance by treating memory from multiple sources as a single addressable block, rather than moving data around between multiple blocks. Software and toolkit support is a differentiator, with associated development effort, maintenance cost, and impact on overall system architecture.

## 4   Conclusions

Machine vision applications in autonomous vehicles use data from a variety of sensors including cameras, radar, and lidar to enable the vehicle to perceive its environment. A variety of processor architectures are available for processing data in these applications. Some tasks are better suited to more general-purpose architectures such as CPUs and GPUs. Others can benefit from specialized processing architectures such as neural processing units,

DSPs, or ASICs. Choosing the right architecture for a given task helps meet safety and performance demands of machine vision applications in autonomous driving. The authors are researching ML accelerator architectures for autonomous driving systems that support L4+ autonomy. They are designing next-generation autonomy compute systems that leverage the power of ML accelerators. Ongoing and future research will provide deeper analysis of specific machine vision use cases using data from a variety of sensors including camera, lidar, and radar.